%% file: main.tex
\documentclass{article}

\usepackage[preprint]{neurips_2026}

\usepackage[utf8]{inputenc}
\usepackage[T1]{fontenc}
\usepackage{newunicodechar}  %
\newunicodechar{ć}{\'c}\newunicodechar{č}{\v{c}}\newunicodechar{ğ}{\u{g}}\newunicodechar{ł}{\l{}}
\newunicodechar{ń}{\'n}\newunicodechar{ž}{\v{z}}
\usepackage{hyperref}
\usepackage{url}
\usepackage{xurl}  %
\usepackage{booktabs}
\usepackage{amsmath,amssymb,amsfonts}
\usepackage{nicefrac}
\usepackage{microtype}
\usepackage{xcolor}
\usepackage{graphicx}
\usepackage{tabularx}
\usepackage{mdframed}
\usepackage{needspace}
\usepackage{placeins}  %

\usepackage[capitalise,nameinlink]{cleveref}

\setcitestyle{authoryear,square,comma}
\hypersetup{colorlinks=true, linkcolor=blue!60!black, citecolor=blue!60!black, urlcolor=blue!60!black}

\newmdenv[backgroundcolor=black!4, linecolor=black!25, linewidth=0.4pt, innerleftmargin=8pt,
  innerrightmargin=8pt, innertopmargin=6pt, innerbottommargin=6pt, skipabove=6pt, skipbelow=6pt]{takeaway}

\newcommand{\file}[1]{\texttt{#1}}
\newcommand{\recordid}[1]{\nolinkurl{#1}}

\newcommand{\mQwen}{\textsc{Qwen3-Coder-30B-A3B}}
\newcommand{\mNemo}{\textsc{Nemotron-Nano-3-30B}}
\newcommand{\mGptoss}{\textsc{gpt-oss-20b}}
\newcommand{\mGemma}{\textsc{Gemma-4-26B-A4B}}
\newcommand{\hCodex}{Codex CLI}
\newcommand{\hOpencode}{OpenCode}
\newcommand{\hMini}{mini-SWE-agent}
\newcommand{\hSweagent}{SWE-agent}
\newcommand{\hDsh}{DeepSeek harness}
\newcommand{\benchmark}{SWE-bench-Live}
\newcommand{\dataseturl}{\url{https://huggingface.co/datasets/pzarzycki/swebenchlive-harness-compact-moe}}
\newcommand{\datasetrevision}{e8789cede466d13625eb2744cfc04ec7c736e574}

\input{numbers}

\title{What Compact MoE Models Reveal\\About Coding Harnesses}
\author{%
  Pawel Zarzycki
}

\begin{document}
\maketitle
\begin{abstract}
\input{sections/abstract}
\end{abstract}
\input{sections/introduction}
\input{sections/related}
\input{sections/design}
\input{sections/interaction}
\input{sections/control}
\input{sections/mechanisms}
\input{sections/discussion}
\input{sections/threats}
\input{sections/conclusion}
\input{sections/statements}

\bibliographystyle{plainnat-etal}
\bibliography{references}

\appendix
\input{sections/appendix}

\end{document}

%% file: numbers.tex
\newcommand{\nCaptured}{8,256}
\newcommand{\clipLo}{0.005}
\newcommand{\clipHi}{0.995}
\newcommand{\gemmaReportTotal}{26B}
\newcommand{\cmBoot}{400}
\newcommand{\nRerun}{1,007}
\newcommand{\coreRerun}{782}
\newcommand{\nTasks}{413}
\newcommand{\nModels}{4}
\newcommand{\nHarnesses}{5}
\newcommand{\nSlots}{8,260}
\newcommand{\nRepos}{234}
\newcommand{\agentDeadline}{3,600}
\newcommand{\coreGradeBudget}{900}
\newcommand{\gemmaGradeBudget}{5,400}
\newcommand{\dshGradeBudget}{21,600}
\newcommand{\deadlineProxy}{3,590}
\newcommand{\latencyMinTokens}{10}
\newcommand{\nCoreSlots}{4,956}
\newcommand{\nValItems}{300}
\newcommand{\nRandItems}{200}
\newcommand{\nPiItems}{40}
\newcommand{\valRunPrec}{0.50}
\newcommand{\valFailEditRecall}{0.64}
\newcommand{\valHoldCls}{0.920}
\newcommand{\valHoldClsW}{0.960}
\newcommand{\valHoldErr}{0.921}
\newcommand{\valPiFirst}{0.875}
\newcommand{\valKappa}{0.957}
\newcommand{\covBoot}{2,000}
\newcommand{\covOriginalQCodMiniN}{297}
\newcommand{\covOriginalQCodMiniBoth}{30}
\newcommand{\covOriginalQCodMiniAonly}{20}
\newcommand{\covOriginalQCodMiniBonly}{22}
\newcommand{\covOriginalQMiniSweN}{332}
\newcommand{\covOriginalQMiniSweBoth}{38}
\newcommand{\covOriginalQMiniSweAonly}{12}
\newcommand{\covOriginalQMiniSweBonly}{43}
\newcommand{\covOriginalNCodOpenN}{346}
\newcommand{\covOriginalNCodOpenAonly}{11}
\newcommand{\covOriginalNCodOpenBonly}{12}
\newcommand{\seqQEditOnePct}{34.7\%}
\newcommand{\seqQEditOneLo}{29.3\%}
\newcommand{\seqQEditOneHi}{40.6\%}
\newcommand{\seqGEditOnePct}{6.1\%}
\newcommand{\seqGEditOneLo}{1.2\%}
\newcommand{\seqGEditOneHi}{11.7\%}
\newcommand{\seqQEditThreePct}{56.5\%}
\newcommand{\seqQEditThreeLo}{50.5\%}
\newcommand{\seqQEditThreeHi}{62.1\%}
\newcommand{\seqGEditThreePct}{24.4\%}
\newcommand{\seqGEditThreeLo}{15.0\%}
\newcommand{\seqGEditThreeHi}{33.3\%}
\newcommand{\seqQEditFivePct}{66.0\%}
\newcommand{\seqQEditFiveLo}{60.2\%}
\newcommand{\seqQEditFiveHi}{71.6\%}
\newcommand{\seqGEditFivePct}{37.8\%}
\newcommand{\seqGEditFiveLo}{27.1\%}
\newcommand{\seqGEditFiveHi}{48.2\%}
\newcommand{\seqQEditTenPct}{73.0\%}
\newcommand{\seqQEditTenLo}{68.0\%}
\newcommand{\seqQEditTenHi}{77.9\%}
\newcommand{\seqGEditTenPct}{65.9\%}
\newcommand{\seqGEditTenLo}{55.4\%}
\newcommand{\seqGEditTenHi}{76.0\%}
\newcommand{\seqQEditEventualPct}{73.7\%}
\newcommand{\seqQEditEventualLo}{68.7\%}
\newcommand{\seqQEditEventualHi}{78.7\%}
\newcommand{\seqGEditEventualPct}{80.5\%}
\newcommand{\seqGEditEventualLo}{71.1\%}
\newcommand{\seqGEditEventualHi}{88.7\%}
\newcommand{\seqQSlots}{285}
\newcommand{\seqQRetryPct}{7.7\%}
\newcommand{\seqQChangedN}{22}
\newcommand{\seqGSlots}{82}
\newcommand{\seqGRetryPct}{91.5\%}
\newcommand{\seqGSameN}{22}
\newcommand{\seqGChangedN}{53}
\newcommand{\seqDiffEditThree}{32.1}
\newcommand{\seqDiffEditThreeLo}{21.5}
\newcommand{\seqDiffEditThreeHi}{43.4}
\newcommand{\seqDiffEditEventual}{-6.8}
\newcommand{\seqDiffEditEventualLo}{-16.5}
\newcommand{\seqDiffEditEventualHi}{4.4}
\newcommand{\seqLaterCalls}{5,996}
\newcommand{\seqBoot}{2,000}
\newcommand{\contractGMatchedSlots}{413}
\newcommand{\contractGSlotsWithCalls}{410}
\newcommand{\pathNCodMiniRawNoExecB}{56}
\newcommand{\pathNCodMiniZeroRequestsB}{55}
\newcommand{\pathNCodMiniRejectedOnlyB}{1}
\newcommand{\coopTasks}{411}
\newcommand{\coopFeatureRows}{4,947}
\newcommand{\coopBehaviorRows}{4,935}
\newcommand{\coopOutcomeRows}{4,943}
\newcommand{\coopClogitLR}{229.43}
\newcommand{\coopClogitDF}{6}
\newcommand{\coopClogitWaldP}{\ensuremath{5.22\times 10^{-21}}}
\newcommand{\coopQMini}{62}
\newcommand{\coopQSwe}{101}
\newcommand{\coopNMini}{88}
\newcommand{\coopNSwe}{79}
\newcommand{\fullAnySuccess}{199}
\newcommand{\fullStrictTasks}{370}
\newcommand{\fullPairedComparisons}{70}
\newcommand{\fullQPairN}{411}
\newcommand{\fullQDeepSuccess}{100}
\newcommand{\fullQSweSuccess}{101}
\newcommand{\fullQPairBoth}{63}
\newcommand{\fullQPairDiscordant}{75}
\newcommand{\fullQPairBothEdit}{59}
\newcommand{\fullQPairBothGrade}{45}
\newcommand{\fullOutcomeRows}{8,237}
\newcommand{\fullBehaviorRows}{8,228}
\newcommand{\fullFeatureRows}{8,247}
\newcommand{\fullGemmaPairN}{411}
\newcommand{\fullGemmaMiniPatch}{345}
\newcommand{\fullGemmaSwePatch}{195}
\newcommand{\fullGemmaMiniSuccess}{94}
\newcommand{\fullGemmaSweSuccess}{49}
\newcommand{\fullNemoPairN}{412}
\newcommand{\fullNemoCodAct}{155}
\newcommand{\fullNemoDshAct}{406}
\newcommand{\fullNemoCodPatch}{47}
\newcommand{\fullNemoDshPatch}{152}
\newcommand{\fullNemoCodSuccess}{13}
\newcommand{\fullNemoDshSuccess}{34}
\newcommand{\fullGemmaMiniPassPatch}{27.2\%}
\newcommand{\fullGemmaSwePassPatch}{25.1\%}
\newcommand{\fullGemmaPassPatchDiff}{2.1}
\newcommand{\fullGemmaPassPatchLo}{-3.4}
\newcommand{\fullGemmaPassPatchHi}{8.3}
\newcommand{\fullFeedbackRows}{2,188}
\newcommand{\fullFeedbackMin}{9}
\newcommand{\fullFeedbackMax}{250}
\newcommand{\fullFeedbackPairN}{95}
\newcommand{\fullFeedbackDshEarly}{42}
\newcommand{\fullFeedbackDshLater}{84}
\newcommand{\fullFeedbackSweEarly}{80}
\newcommand{\fullFeedbackSweLater}{92}
\newcommand{\fullFeedbackEditorN}{61}
\newcommand{\fullFeedbackEditorDshEarly}{23}
\newcommand{\fullFeedbackEditorDshLater}{53}
\newcommand{\fullFeedbackEditorSweEarly}{51}
\newcommand{\fullFeedbackEditorSweLater}{59}
\newcommand{\fullFeedbackOriginalN}{84}
\newcommand{\fullFeedbackOriginalDshEarly}{38}
\newcommand{\fullFeedbackOriginalSweEarly}{73}
\newcommand{\fullFeedbackOriginalDshLater}{73}
\newcommand{\fullFeedbackOriginalSweLater}{82}

%% file: sections/abstract.tex
Why does the same compact model solve a repository issue in one coding harness but fail in another? Similar scores can conceal different solved issues for the same coding model. We study model--harness cooperation by evaluating four compact open-weight mixture-of-experts models (\mQwen{}, \mNemo{}, \mGptoss{} and \mGemma{}) on \nTasks{} \benchmark{} MultiLang issues using \hCodex{}, \hOpencode{}, \hMini{}, \hSweagent{} and the \hDsh{}. We describe all configurations as run, with a common-cohort check across three models and four harnesses. Every model and every harness contributes exclusive successes, including on the subset with eligible outcomes and no grader timeouts across all configurations.

On \fullQPairN{} jointly eligible issues, Qwen solves \fullQDeepSuccess{} in the \hDsh{} and \fullQSweSuccess{} in \hSweagent{}, with \fullQPairBoth{} shared successes. On \fullQPairBothEdit{} of the \fullQPairDiscordant{} issues with different outcomes, both traces record an edit attempt. The \hMini{} and \hSweagent{} comparisons extend this pattern to Nemotron and Gemma, while both traces seldom record edits when gpt-oss succeeds under only one harness. Gemma acts on every matched issue in both harnesses, with substantially more retained patches and successes under \hMini{}.

Across the full grid, planning share varies most across harnesses, while editing and reading shares vary most across particular model--harness pairings. Paired traces show differences in which files agents edit, what changes they retain and how they continue after tool refusals or failed edits. We publish \nSlots{} retained results and \nCaptured{} captured trajectories.

%% file: sections/introduction.tex
\section{Introduction}\label{sec:introduction}

A repository issue is only the starting point for a coding agent. The model has to interpret the issue, decide what to inspect or change, and revise its work in light of what it observes. The harness constructs the model's inputs, provides and executes actions, returns observations, and determines when the interaction continues or ends. We use \emph{cooperation} to describe how a model and a configured harness work together through this loop.

We ran four compact open-weight mixture-of-experts models through five coding harnesses on the same \nTasks{} \benchmark{} issues in every configuration. Each model contributes successes on issues that no other model solves in any harness, and each harness contributes successes on issues that no other harness solves with any model. Both findings hold on the \fullStrictTasks{} issues with eligible outcomes and no grader timeouts across the full grid. For Qwen, the \hDsh{} and \hSweagent{} solve almost exactly the same number of issues eligible in both configurations, yet each succeeds on dozens of issues the other misses. Many of those differences persist through editing and delivery of a graded patch.

Crossed studies already show that model and harness rankings depend on the pairing, and that successful issue sets can differ \citep{yao2026harnessbenchevaluatingagentharnesses,li2026findingrightfitmodelharness,hu2026modeldemystifyingharnesseffects}. We examine where those differences appear along the execution. Matched runs distinguish issues where an agent never acts from those where both agents edit and still earn different grades. The recorded interactions let us examine what the agents changed, what they did after a refusal or a failed edit, and what reached grading. We use the three-model, four-harness block that shares a collection cohort and serving route as a sensitivity analysis.

\paragraph{Contributions.}
\begin{itemize}
  \item \textbf{Exclusive and shared issue coverage.} We measure which issues each model and harness contributes to the observed success set, and compare gains and losses within pairs of configurations.
  \item \textbf{Where paired executions diverge.} We distinguish action entry, editing, retained patches and success. We compare workflow across the grid and examine paired traces for differences in what agents change and how they continue after setbacks.
  \item \textbf{A crossed evaluation and public execution corpus.} We release the retained results and captured trajectories, linking outcomes to recorded inputs, tool interfaces, requests, feedback and patches.
\end{itemize}

%% file: sections/related.tex
\section{Related work}\label{sec:related}

\paragraph{Crossed harness evaluations.}
Scaffold choice changes agent accuracy and cost \citep{kapoor2025holisticagentleaderboardmissing}. Observational SWE-bench comparisons show substantial within-model scaffold ranges and model--scaffold interactions \citep{liu2026codingagentsconvergedswebench}. Controlled crossed studies report pairing-dependent rankings, complementary successes, and cases of feedback use and incomplete delivery on workplace and terminal tasks \citep{yao2026harnessbenchevaluatingagentharnesses,li2026findingrightfitmodelharness}. In repository tasks, harness complexity interacts with model strength and with the type of work required \citep{hu2026modeldemystifyingharnesseffects}. A controlled GAIA comparison examined whether model strength predicts scaffold sensitivity \citep{starace2026scaffoldeffectsgaiacontrolled}. Our study adds a common-task compact-MoE evaluation of repository repair that links shared and exclusive successes to whether both agents edit, what they change, and what they do after the same verified refusal.

\paragraph{Execution trajectories.}
Trace studies identify workflow phases, action motifs and sequences through which failures propagate \citep{bouzenia2025understandingsoftwareengineeringagents,liu2026processcentricanalysisagenticsoftware,zhao2026failureprocessanatomycli}. Observational work finds that shared models predict task-level agreement \citep{mehtiyev2026resolutionratesbehavioraldrivers} and that the relation between behavioural measures and resolution varies across frameworks \citep{ma2026signaldifferentsemanticscrossframework}. \citet{raj2026modelharnessinteractioncentrictaxonomy} distinguish model, harness and interface events in individual failures. Our same-task execution comparisons cover executable actions, edit attempts, source changes and benchmark outcomes. Both successful and unsuccessful paths inform the comparison.

\paragraph{Action interfaces and continuation.}
Tool descriptions affect invocation \citep{johnson2025naturallanguagetoolsnatural,somma2026remarkableeffectivenessprovidingai}, and training trajectories expose models to scaffold-specific interfaces \citep{yang2025swesmithscalingdatasoftware,pan2025trainingsoftwareengineeringagents}. Component ablations connect planning prompts with fewer stops before editing, and the same work reports a Nemotron model using familiar tool patterns in a shell-only harness \citep{fan2026empiricalstudyharnessdesign}. \citet{mavrin2026harmonygptoss} document Harmony integration and unavailable patch requests for gpt-oss. Recorded inputs, exact tool menus and linked feedback let us identify which actions were available and follow the work that came next. Harness versions and context management also affect outcomes \citep{sghaier2026dontblamelargelanguage,lewis2026modeldifferentharnessdifferent}, so our comparisons concern the integrated configurations.

%% file: sections/design.tex
\section{Study and measurements}\label{sec:design}

\paragraph{Common tasks across the full grid.}
We evaluated \nModels{} compact open-weight mixture-of-experts models in \nHarnesses{} harness families on the same \nTasks{} \benchmark{} MultiLang issues \citep{zhang2025swebenchgoeslive,li2026repolaunchautomatingbuildmanagement}. The frozen Linux task set spans \nRepos{} repositories. We call each model--harness pairing a \emph{lane} and its retained observation on an assigned task a \emph{slot}. The full grid contains \nSlots{} assigned slots and \nCaptured{} captured trajectories. The common-cohort block crosses \mQwen{}, \mNemo{} and \mGptoss{} with \hCodex{}, \hOpencode{}, \hMini{} and \hSweagent{} in one cohort and serving route (\nCoreSlots{} assigned slots). We report full-grid comparisons of retained configurations and repeat key comparisons within that block to see whether they also hold under shared study conditions. The additional \hDsh{} and \mGemma{} runs span separate cohorts, serving routes and grading budgets (\cref{sec:configuration_details}).

\paragraph{Configured harnesses.}
All models were served through Amazon Bedrock. A harness here includes its pinned prompts, exposed tools, adapter, serving integration and continuation rules. The harnesses addressed models by a generic name, so model-specific branches were inactive. \hMini{} accepts a single fenced shell command in the visible reply. The other harnesses use function calls. Mini and SWE-agent require explicit submission, while Codex, OpenCode and the DeepSeek harness end on a reply without a tool call, subject to the first-turn requirement in one DeepSeek variant. All runs had an agent wall-clock budget of \agentDeadline{} seconds and no internet. The integrations determine which proposals can become actions, which feedback reaches the next call, and which replies end the task.

\paragraph{Outcomes and trace populations.}
A retained patch succeeds when the evaluator observes at least one listed fail-to-pass test passing and no listed test failing. Deadline-ended agents are scored on the changes they leave. The full-grid outcome sample has \fullOutcomeRows{} eligible rows. It excludes infrastructure or benchmark outcomes without usable grades and counts grader timeouts as nonsuccess. Joining outcomes to behaviour also requires captured evidence and agreement on selected-attempt identity and outcome, yielding \fullBehaviorRows{} rows. Describing trace features alone uses the matched capture base of \fullFeatureRows{} rows. \Cref{sec:coverage_details} specifies each denominator and the retained-attempt selection rule.

\paragraph{Actions, edits and progress.}
A deterministic classifier maps shell commands and named tools to shared classes such as reading, searching, editing, planning, testing and building. Protocol failures and unavailable-tool requests are coded separately from executable work. An observed edit attempt can change a scratch file, fail to change a file, or affect source without entering a patch. For paired paths we count recorded edit attempts on each side. For progress profiles we use nested milestones for acting, editing, patch and success, with edit and action inferred from a retained patch when required. A patch can enter the progress profile even when grading timed out. Workflow shares describe the division of executable action records. \Cref{sec:measurement_details} gives the edit conventions, treatment of compound actions and validation results. In held-out actions, classifier agreement with blind model-coder consensus was \valHoldCls{} for class (\valHoldClsW{} after weighting to the population class mix) and \valHoldErr{} for the error flag, excluding disputed items. Agreement with the author's independent first-pass class labels on a separate random sample was \valPiFirst{}.

\paragraph{Comparisons and uncertainty.}
Paired coverage counts shared and exclusive successes on tasks eligible in the compared configurations. Within the common-cohort block, a task-conditional logit compares an additive success model with one allowing model--harness interactions. For trace features, a finite-grid decomposition separates model, harness and interaction variation in task-adjusted lane averages. We also measure when editing resumes after recorded edit errors across the grid and reconstruct effective model inputs for the shared Codex refusal. We bootstrap at task level, keeping paired observations together. \Cref{sec:analysis_methods,sec:feedback_methods,sec:sequence_methods} gives the estimands and uncertainty.

%% file: sections/interaction.tex
\section{Which pairings succeed on which issues?}\label{sec:interaction}

\paragraph{Every model and harness contributes exclusive successes.}
Each model solves issues that none of the other models solves under any harness. Each harness also solves issues that none of the other harnesses solves with any model. Both patterns hold among the \fullStrictTasks{} issues with eligible outcomes and no grader timeouts in all configurations (\cref{tab:full_exclusivity}). Across \nTasks{} assigned issues per configuration (\cref{tab:assigned_grid}), at least one configuration succeeds on \fullAnySuccess{} issues. This observed union combines the results of running all configurations on each issue, with multiple executions per issue.

\begin{table}[t]
  \centering
  \caption{Retained successes out of \nTasks{} assigned benchmark issues per configuration. The full grid includes configurations collected under different cohorts and serving routes.}
  \label{tab:assigned_grid}
  \small\input{tables/a_assigned_grid}
\end{table}

\begin{table}[t]
  \centering
  \caption{Issues solved exclusively by one model or one harness. For a model, other models must have no success under any harness. For a harness, other harnesses must have no success with any model. The strict cohort requires eligible outcomes and no grader timeouts across all configurations.}
  \label{tab:full_exclusivity}
  \small\input{tables/full_exclusivity}
\end{table}

\paragraph{Nearly equal totals can cover different issues.}
On the \fullQPairN{} issues eligible in both Qwen configurations, the \hDsh{} solves \fullQDeepSuccess{} and \hSweagent{} solves \fullQSweSuccess{}. The two share only \fullQPairBoth{} successes, so a one-issue difference in totals accompanies \fullQPairDiscordant{} different outcomes. Among these discordant issues, both traces record an executable edit attempt for \fullQPairBothEdit{} and both configurations submit a nonempty patch with a decisive grade for \fullQPairBothGrade{}. These paired observations show that the difference extends beyond action entry and edit attempts, through to the material submitted for grading. The two configurations were collected in different cohorts.

\paragraph{On most discordant issues, both runs attempt edits for three models.}
Under \hMini{} and \hSweagent{}, each model gains successes on some issues and loses them on others (\cref{tab:full_pair_paths}). Among issues with different outcomes, both traces record an edit attempt on most issues for Qwen, Nemotron and Gemma. For gpt-oss, both traces rarely record edits on such issues. The recorded action and patch milestones help distinguish the kinds of divergence across these pairs (\cref{sec:control}).

\begin{table}[t]
  \centering
  \caption{Paired outcomes under \hMini{} and \hSweagent{}, using each model's common eligible issues. Discordant issues are solved under exactly one harness. The edit column counts both-side edit attempts among discordant issues with two matched traces. Both graded counts discordant issues with two nonempty patches that received success or failure grades.}
  \label{tab:full_pair_paths}
  \small\input{tables/full_pair_paths}
\end{table}

\paragraph{Rank reversal within the shared cohort.}
Requiring eligibility in all four Qwen and Nemotron configurations under these two harnesses gives a single set of \coopTasks{} issues. Under \hMini{}, Nemotron leads Qwen with \coopNMini{} versus \coopQMini{} successes. Under \hSweagent{}, Qwen leads Nemotron with \coopQSwe{} versus \coopNSwe{}. This rank reversal uses the same issues for both models. The counts in \cref{tab:full_pair_paths} use each model's own pair-eligible cohort.

As a separate check in the three-model, four-harness block with a common collection cohort and serving route, a task-conditional logit supports a model--harness interaction (likelihood ratio \coopClogitLR{} on \coopClogitDF{} degrees of freedom and Wald \(p=\coopClogitWaldP{}\) using task-bootstrap covariance). It persists when the three cells with severe action-interface difficulties are omitted (\cref{tab:outcome_models}). Additional paired coverage counts and preserved-original sensitivity results are reported in \cref{sec:task_coverage}.

%% file: tables/a_assigned_grid.tex
%
\begin{tabular}{@{}lrrrrr@{}}
\toprule
Model & Codex CLI & OpenCode & mini-SWE-agent & SWE-agent & DeepSeek (ext.) \\
\midrule
Qwen & 61 & 67 & 63 & 102 & 101 \\
Nemotron & 14 & 14 & 89 & 79 & 34 \\
gpt-oss & 66 & 20 & 2 & 59 & 40 \\
Gemma-4 (ext.) & 66 & 57 & 94 & 49 & 61 \\
\bottomrule
\end{tabular}

%% file: tables/full_exclusivity.tex
%
\begin{tabular}{@{}lr@{\qquad}lr@{}}
\toprule
Model & Exclusive issues & Harness & Exclusive issues \\
\midrule
Qwen & 16 & Codex & 7 \\
Nemotron & 4 & DeepSeek & 5 \\
gpt-oss & 12 & mini & 5 \\
Gemma & 12 & OpenCode & 3 \\
 &  & SWE-agent & 17 \\
\bottomrule
\end{tabular}

%% file: tables/full_pair_paths.tex
%
\begin{tabular}{@{}lrrrrrr@{}}
\toprule
Model & Tasks & Mini only & SWE only & Discordant & Both edit / matched & Both graded \\
\midrule
Qwen & 412 & 18 & 57 & 75 & 68/75 & 30 \\
Nemotron & 412 & 37 & 28 & 65 & 58/65 & 37 \\
gpt-oss & 413 & 1 & 58 & 59 & 5/59 & 0 \\
Gemma & 412 & 61 & 16 & 77 & 49/77 & 21 \\
\bottomrule
\end{tabular}

%% file: sections/control.tex
\section{From actions to repository changes}\label{sec:control}

\begin{figure}[t]
  \centering
  \includegraphics[width=\linewidth]{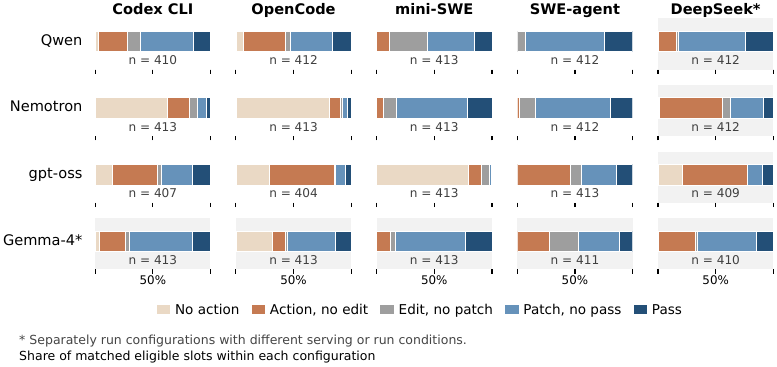}
  \caption{Furthest progress milestone across \fullBehaviorRows{} eligible slots with matched captures in the full grid. All percentages use the matched denominator for each configuration. Shading marks the separately run configurations. A retained patch can imply an action or edit when recorded evidence is missing, and the patch milestone includes grader timeouts. \Cref{sec:measurement_details} defines the milestones.}
  \label{fig:progress_profiles}
\end{figure}

\paragraph{Progress separates action entry from leaving a patch.}
\Cref{fig:progress_profiles} shows the furthest milestone recorded across the full grid. For Gemma under mini and SWE-agent, both configurations acted on all \fullGemmaPairN{} matched tasks. Mini left \fullGemmaMiniPatch{} patches and passed \fullGemmaMiniSuccess{} tasks, while SWE-agent left \fullGemmaSwePatch{} patches and passed \fullGemmaSweSuccess{} tasks. The pass shares among the patches each left were \fullGemmaMiniPassPatch{} and \fullGemmaSwePassPatch{}, over different sets of tasks (difference \fullGemmaPassPatchDiff{} percentage points, 95\% CI \fullGemmaPassPatchLo{} to \fullGemmaPassPatchHi{}). Gemma's gap in observed passes accompanies a substantial difference in how often the agents leave a patch.

For Nemotron under the DeepSeek harness and Codex, the difference appears at action entry. On \fullNemoPairN{} matched tasks, we observe executable actions on \fullNemoDshAct{} and \fullNemoCodAct{} tasks, respectively. Their executions left \fullNemoDshPatch{} and \fullNemoCodPatch{} patches and passed \fullNemoDshSuccess{} and \fullNemoCodSuccess{} tasks. This contrast spans recording, cohort and configuration differences. A deeper audit of Nemotron under Codex and mini found both replies without structured requests and requests explicitly rejected by the harness (\cref{sec:path_details}).

\paragraph{Task framing and action form affect the next step.}
On a samtools feature request phrased as a question, gpt-oss under Codex replies with usage advice and the interaction ends. The SWE-agent request appends an explicit instruction to implement repository changes, while Codex's system message directs autonomous issue resolution. Under mini, an action requires a shell command in the visible reply. Some gpt-oss calls record reasoning with no visible reply, illustrating how what reaches the harness can constrain action entry (\cref{sec:context_audit}).

\begin{figure}[t]
  \centering
  \includegraphics[width=\linewidth]{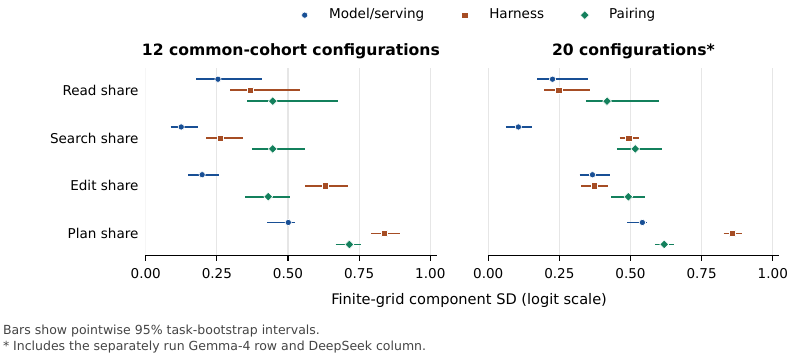}
  \caption{Task-adjusted workflow variation in the full grid and common-cohort block. Points give finite-grid SDs associated with model and serving configuration, harness, and their pairing. Bars give 95\% pointwise task-bootstrap intervals. All plotted features use the logit scale. Compare component magnitudes within a feature.}
  \label{fig:behavioral_variation}
\end{figure}

\paragraph{The full grid changes which components are largest.}
The harness component is largest for planning share in both the common-cohort block and the full grid. The harness component is also largest for edit share in the common-cohort block, and the pairing component is largest in the full grid. For both edit and read shares, the full-grid paired intervals separate the pairing component from the harness and model components. For searching, the harness and pairing components are close and the paired interval for their difference spans zero. These estimates describe how recorded action shares vary across the observed configurations, including differences in the granularity of available tools (\cref{fig:behavioral_variation,sec:control_details}).

\paragraph{Editing spans different kinds of work.}
Three same-task Qwen comparisons under mini and SWE-agent show what an edit milestone can include. In TanStack, mini writes explanatory files and leaves an empty patch. SWE-agent changes router source and leaves a successful patch despite unsuccessful package-test attempts. In Kepler, both agents change VM detection code. The SWE-agent execution retains a pattern after its own Python regex probes report mismatches on legacy paths, while mini leaves a different rewrite and succeeds. In Valkey, both agents leave different changes to the same C source file and both succeed despite unsuccessful local checks. \Cref{sec:late_path_cases} gives the case selection and locators.

%% file: sections/mechanisms.tex
\section{Feedback, continuation and completion}\label{sec:mechanisms}

\begin{figure}[t]
  \centering
  \includegraphics[width=\linewidth]{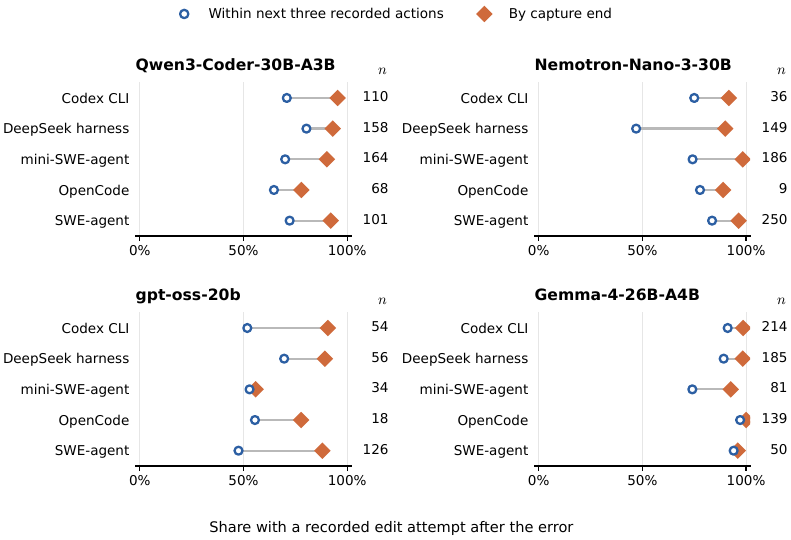}
  \caption{Recorded edit attempts after the first eligible edit error in each selected run. Each pairing's denominator \(n\) includes all runs with an eligible error, including captures with fewer than three subsequent actions. The markers show observed fractions among these runs. Shell exits and editor errors differ in what they record, and the longer window ends with the available capture.}
  \label{fig:feedback_continuation}
\end{figure}

\paragraph{Editing continued at different points in the recorded trajectories.}
In all model--harness pairings, more runs recorded another edit attempt by capture end than within the next three recorded actions after the first eligible edit error (\cref{fig:feedback_continuation}). The analysis covers \fullFeedbackRows{} matched runs, with \fullFeedbackMin{} to \fullFeedbackMax{} per pairing.
On \fullFeedbackPairN{} tasks where Nemotron had an eligible error under both \hDsh{} and \hSweagent{}, the next three recorded actions included an edit attempt on \fullFeedbackDshEarly{} tasks under \hDsh{} and \fullFeedbackSweEarly{} under \hSweagent{}. By capture end the counts were \fullFeedbackDshLater{} and \fullFeedbackSweLater{}. Both traces recorded at least three subsequent actions on every shared task. The gap also narrowed when both errors came from \texttt{str\_replace\_editor} and only edits in a different recorded model call counted (\cref{sec:feedback_methods}). These comparisons describe the recorded order of events. Error content and remaining opportunities for editing vary across pairings, and whether a model received the error is unverified.

\paragraph{The same refusal can lead to different next actions.}
Codex instructs the model to use \texttt{apply\_patch} even when its offered menu omits that tool. We follow a common first refusal for \seqQSlots{} Qwen and \seqGSlots{} gpt-oss executions sharing a normalised prompt, an exact menu and a refusal text verified in the next effective model input. Across \seqLaterCalls{} later recorded model calls the tool stays absent, the refusal remains visible and each slot's system and developer instructions stay unchanged. Gpt-oss requests the unavailable tool again in \seqGRetryPct{} of slots, compared with \seqQRetryPct{} for Qwen. Payloads sometimes change across retries. The request still names an unavailable action, so even changed arguments can leave repository work stalled.

In the MetaMask comparison, Qwen under Codex receives the same type of refusal and goes on to read, search and use \texttt{sed} through the offered shell tool. The edit and the submitted patch are recorded, and both Codex and mini succeed on that issue. The shell channel offered another way to make a change within the same execution. \Cref{sec:recovery_case} gives the call sequence and patch locator.

\begin{figure}[t]
  \centering
  \includegraphics[width=\linewidth]{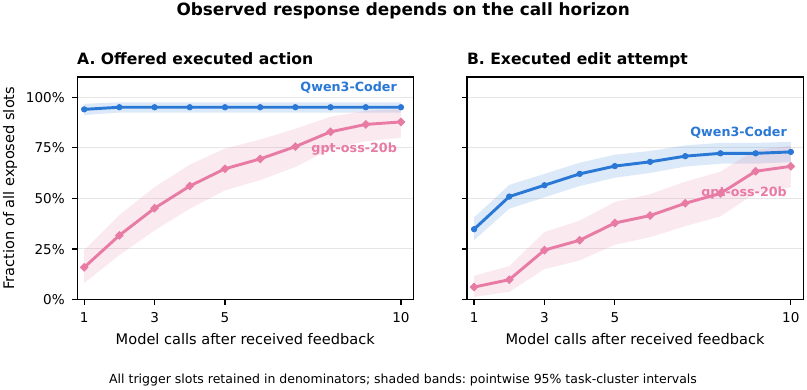}
  \caption{Recorded responses over model-call horizons after a common verified refusal. Curves show the fraction of trigger slots reaching (A) an offered executable action or (B) an executed edit attempt, with 95\% pointwise task-cluster bootstrap bands. The first call receiving the refusal is the first decision opportunity. All trigger slots stay in the denominator, including captures that end before a horizon. The text reports the remaining-capture endpoints.}
  \label{fig:refusal_sequences}
\end{figure}

\paragraph{The editing contrast changes with the horizon.}
Within three model calls of receiving the refusal, \seqQEditThreePct{} of Qwen slots and \seqGEditThreePct{} of gpt-oss slots record an executed edit attempt (Qwen minus gpt-oss \seqDiffEditThree{} percentage points, 95\% CI \seqDiffEditThreeLo{} to \seqDiffEditThreeHi{}). Across the remaining captures, the fractions are \seqQEditEventualPct{} and \seqGEditEventualPct{}, respectively (difference \seqDiffEditEventual{} points, 95\% CI \seqDiffEditEventualLo{} to \seqDiffEditEventualHi{}). The early Qwen advantage gives way to an imprecise remaining-capture difference (\cref{fig:refusal_sequences}). Persistent retrying can coexist with later editing. The remaining-capture endpoint spans variable lengths, and this comparison conditions on a refusal produced within each pairing. The intervals describe the selected exposures, with a separate shared-task check in \cref{sec:sequence_methods}.

\paragraph{Completion depends on the remaining interaction.}
A reply without a tool call can end Codex or OpenCode work, while the two explicit-submit harnesses offer a distinct way to declare the task complete. In both forms, a run can reach its agent deadline before further work, leaving the benchmark to grade the patch that remains. Local feedback can also run out of useful evidence when the required command is missing, as in the Valkey and TanStack cases. These paths separate another action, another chance to interpret feedback and delivery of a graded patch. Timing and continuation vary with the serving configuration as well as with the harness loop (\cref{tab:a_route}).

%% file: sections/discussion.tex
\section{Discussion}\label{sec:discussion}

\paragraph{Issue coverage depends on the configured pairing.}
The union of observed successes shows how much issue coverage is distributed across the grid. Model and harness ranks summarise how often configurations succeed. The paired outcomes show which issues they gain or lose, and the size of those changes can substantially exceed the difference in totals. The observed union requires multiple executions per issue. Comparing strategies under the same budget would require measuring how they allocate those attempts and select a patch.

\paragraph{The stage where executions diverge varies by pairing.}
A model must interpret the task and express its next step in an action format the harness can use. The samtools exchange shows how a question-shaped issue can draw an advisory reply even under a directive to resolve the issue. The gpt-oss/mini records show a related demand at action entry: a proposed next step has to appear in the output channel the harness consumes. For Gemma under mini and SWE-agent, both executions act on every issue in the matched cohort, while the number of retained patches differs sharply. Among runs that leave patches, the two success shares are similar. Paired Qwen cases show how runs that both edit can make different changes, act differently after local checks and leave different work for grading.

\paragraph{Workflow reflects the interface and how the model uses it.}
Planning share varies most across harnesses even when the full grid is used. Editing share varies most across particular pairings in the full grid, while the harness component is largest in the shared-cohort block. Reading also varies most across pairings in the full grid, and searching shows substantial variation both across harnesses and across pairings. The recorded action shares reflect the granularity of the available tools as well as decisions about what to do with them.

\paragraph{Continuation depends on available actions and time.}
The Codex refusal records verify common feedback in effective model inputs and show how an available shell route can let work continue. Qwen's fallback in MetaMask illustrates how the harness and model can navigate a refusal through another supported action. The cross-harness edit-failure comparisons describe the order of recorded events. An edit may follow a failed result even where its receipt in the effective model input is unverified. Different action interfaces, completion rules and serving routes shape the observed opportunities to continue. Isolating the effect of any one property would require holding the rest of the configuration fixed.

%% file: sections/threats.tex
\section{Validity and limitations}\label{sec:threats}

\paragraph{Executions and configurations.}
Each slot represents one selected execution. The uncertainty from selecting an attempt and the unobserved variation across repeated attempts are distinct from the task-bootstrap intervals (\cref{sec:coverage_details}). The common-cohort block holds the cohort and serving route fixed. Prompts, tools, adapters and stopping rules still change together across harnesses, and model configurations include their parser and response representation. Full-grid contrasts can span separate cohorts, serving routes and grading budgets. The study population supports comparisons among these compact MoE configurations. A comparison of model scales or architectures would require additional populations.

\paragraph{Measurement and conditioning.}
Classifier validation primarily uses model-coder consensus and a smaller independent author first pass. Result formats vary across harnesses, and edit-effect detection is imperfect. Action shares reflect recording granularity, the progress profiles allow patch-implied events, and some captures lack call boundaries. Route and harness logging differences can also change which actions appear in the capture. The contextual cases are model-assisted close readings. Masked packets and outcome-known paired cases answer different questions. Action shares, reached milestones, refusal exposure and successful patch comparisons can condition on events that the pairing itself influences. These comparisons describe recorded behaviour in those subsets. Pointwise intervals do not adjust for selecting multiple features, horizons or illustrative pairs.

\paragraph{Task population and grading.}
Success applies to the frozen Linux benchmark and its test predicate. Missing local tools limit what agents can verify, and training contamination remains possible. The public records support inspection and independent analysis of the runs. Capture completeness and private measurement rules limit exact reproduction of every derived estimate.

%% file: sections/conclusion.tex
\section{Conclusion}\label{sec:conclusion}

Every model and every harness contributes exclusive successes on this benchmark, including among issues with eligible outcomes and no grader timeouts in every configuration. Nearly equal success totals for Qwen in the \hDsh{} and \hSweagent{} cover substantially different issues. On many of those issues both runs attempt edits, and many yield two graded patches with different outcomes.

Some large outcome gaps come with large differences in action entry, as with gpt-oss in \hMini{}. Other differences emerge during repository work or at submission, as with Gemma in \hMini{} and \hSweagent{}. Paired traces show which files agents change, what they leave in the final patch and which routes they take after tool refusals or failed edits. The released trajectories preserve inputs, actions, tool results and patches so those paths can be examined alongside benchmark grades.

%% file: sections/statements.tex
\section{Data availability}\label{sec:statements}

The benchmark results, task metadata, patches and captured trajectories are public at \dataseturl{}, revision \texttt{\datasetrevision}. We describe the record layout, provenance and access in \cref{sec:data_card}. The analysis and classifier implementation remain private.

%% file: sections/appendix.tex
\section{Data, configurations and analysis populations}\label{sec:appendix}

\subsection{Public records}\label{sec:data_card}
The study records are available at \dataseturl{}, revision \recordid{e8789cede466d13625eb2744cfc04ec7c736e574}. The published file digests match the frozen study export \texttt{final-20261019}. The \file{final-resultset} folder contains structured results, a flat result projection, the task manifest, a patch index, deduplicated patch bodies, documentation and checksums. The \file{agent-action-trajectories} folder contains the selected-slot index, deduplicated conversations, call inputs and outputs, tool schemas and coverage metadata. The dataset viewer exposes the index. Full conversation payloads are in \file{conversations.jsonl.xz}; the split named \texttt{test} denotes the benchmark records.

To read a trajectory, resolve the selected-slot index's \texttt{conversation\_id} to its payload. Each call references messages and can override fields in its own input view. Apply those overrides, including nulls that remove a field. A null input indicates missing context. Join to the retained result by \texttt{unit\_id} and confirm both selection kind and outcome. To find a retained patch, resolve that result's patch identity through the patch index and archive. Explicit markers record redactions of credentials, binary omissions and reference-only patch bodies. Capture omissions and source-time truncation limit the evidence available. The dataset card describes source-content terms. The analysis code, derived feature tables and individual validation labels are private.

The index covers \nSlots{} assigned slots and contains \nCaptured{} captured trajectories and four empty exports. Three empty exports are startup failures and one has recorded requests but a missing collected harness log. Captures combine adapter observations with native harness records. A capture can lack a main-call boundary or contain a partial response. The metadata mark source and context diagnostics. Recorded call inputs describe the harness-to-adapter boundary.

\subsection{Configurations and grading}\label{sec:configuration_details}
The common-cohort \texttt{core-r7} block uses Amazon Bedrock Runtime through LiteLLM. The other cohorts are \texttt{dsh-r6} and \texttt{gemma-r1}. Gemma uses Amazon Bedrock Mantle's OpenAI-compatible route without parallel calls. Some gpt-oss executions in the DeepSeek cohort use Mantle or buffered non-streaming Runtime. \Cref{tab:models_harnesses} describes the models and action interfaces. The harness pins are \hMini{} \texttt{v2.4.6 (a83fcae8)}, \hSweagent{} \texttt{v1.1.0 (0f3acafa)}, \hCodex{} \texttt{0.147.0}, \hOpencode{} \texttt{opencode-ai 1.18.15}, and \hDsh{} \texttt{@deepseek-ai/dsh 0.1.0-rc.7 (99f6f02f)}. The first three models use the \texttt{required-initial-tool-v1} DeepSeek variant and Gemma uses \texttt{locked-default}. The \texttt{unit} model identifier activates the generic integration paths. Qwen is code-specialised, Nemotron has training exposure to OpenHands, SWE-agent and mini-SWE-agent trajectories \citep{nvidia2025nemotron3nanoopen}, gpt-oss uses Harmony channels \citep{openai2025gptoss120bgptoss20bmodel}, and Gemma has its own tool-call format \citep{gemmateam2026gemma4technicalreport}.

\begin{table}[!htbp]
  \centering
  \caption{Study models and configured action interfaces. Counts follow vendor conventions. Nemotron's active count excludes embeddings. Gemma's technical report gives \gemmaReportTotal{} total parameters.}
  \label{tab:models_harnesses}
  \footnotesize
\input{tables/t1_models}\\[4pt]
\input{tables/t1_harnesses}
\end{table}

The Linux task manifest uses \href{https://huggingface.co/datasets/SWE-bench-Live/MultiLang/tree/3430730b50bba3ad11b40ca9ba5b224f4034ce1a}{\benchmark{} MultiLang revision \texttt{3430730b}}. Its main languages are TypeScript, JavaScript, Go, Java and Rust. The final evaluator is based on \texttt{v1.0-multi-language-multi-os-benchmarking (c5ea7e48)} with RepoLaunch \texttt{ff359c6e}. The public \file{grader-auxiliary.md} describes the configured evaluation and evaluator qualifications. Success requires at least one \texttt{FAIL\_TO\_PASS} test to pass, with no observed \texttt{FAIL\_TO\_PASS} or \texttt{PASS\_TO\_PASS} test failing. The base per-slot grading budgets are \coreGradeBudget{} seconds for the common-cohort block, \gemmaGradeBudget{} for Gemma and \dshGradeBudget{} for the DeepSeek cohort. Some common-cohort grades use extended budgets documented in the public export. Agent and grading budgets refer to different processes. \Cref{tab:a_outcomes} gives the retained outcome categories.

\begin{table}[!htbp]
  \centering
  \caption{Retained outcome categories by cohort. DSH denotes \hDsh{}.}
  \label{tab:a_outcomes}
  \small\input{tables/a_outcomes}
\end{table}

\subsection{Analysis populations}\label{sec:coverage_details}
We use one selected execution per assigned slot. The full grid includes \nRerun{} replacement executions, of which \coreRerun{} are in the \nCoreSlots{} common-cohort slots. Most replacements followed failed executions, and some followed completed executions. Grades for the executions these replaced are unavailable. The selection mechanism may therefore affect the observed estimates. Analyses restricted to \texttt{preserved\_original} selections describe that subset of the task--configuration grid.

The outcome sample retains success, failure, empty patch and grader timeout. It excludes provider errors, startup failures, benchmark defects and missing grades, and counts timeouts as nonsuccesses. The full grid has \fullOutcomeRows{} eligible outcomes and the common-cohort block has \coopOutcomeRows{}. The progress and paired-behaviour sample additionally requires a captured trajectory, agreement of selection kind and outcome between the two records, and a defined executable-action count. This yields \fullBehaviorRows{} rows across the full grid and \coopBehaviorRows{} in the common-cohort block. Paired outcome and trace samples intersect tasks across the compared lanes. The workflow feature analysis starts from \fullFeatureRows{} captured slots with matching selection and outcome records, including slots without eligible grades. The corresponding common-cohort base contains \coopFeatureRows{} slots. Each feature uses the rows on which it is defined. The exact-call analyses additionally require the input and output evidence relevant to each endpoint.

\section{Action and workflow measurement}\label{sec:measurement_details}

\paragraph{Actions and execution.}
We build action records from observed assistant outputs and native records. Input-only replays of assistant messages do not add actions. Where an adapter action view exists we use it, and retain native-only actions where those supply the sole action view. Tool results are associated by call identifier where unique. Where a serving route reuses identifiers, we associate them by position in the call window. The classifier recognises reading, searching, exploration, editing, testing, building, running, installation, file operations, version control, planning, delegation, web access, submission, no-op and other actions. Protocol failures and unavailable tools are separate classes. The broad coding uses recorded menus at conversation level and in some cases a lane-level imputed menu. Exact tool availability in the refusal analysis always comes from the call's recorded schema.

In the coded data, an executable action is a record whose class is neither protocol nor unoffered. The label rests on the recorded request and result rather than direct process monitoring. In shell calls, the parser handles wrappers, pipes, heredocs and compound commands and preserves the component intents. Shares assign each executable action one primary class, prioritising an edit over a test or build, and a test or build over reading or searching. Dedicated-tool calls and compound shell calls can therefore represent the same work with different counts. A result-text flag marks errors, and an edit-outcome flag marks a confirmed effect, an explicit failure or an unknown effect. An edit attempt covers all three flags. First-reply action denotes structured tool requests or one command block in mini's visible output.

\paragraph{Progress and other feature definitions.}
For the progress profiles, a patch is a retained nonempty submission recorded as success, failure or grader timeout. An edited slot has an executable edit with a confirmed or unknown effect, or a patch that implies an edit. An acted slot has an executable action or the evidence required by a later milestone. All milestone shares use the full matched lane denominator. These milestones are nested summaries. The paired paths instead use recorded edit attempts, including explicit failed edits, and keep patch-implied events separate.

Across the matched capture base, the patch indicator is one for success or failure and zero for the other recorded outcomes. The first-edit position is the recorded action index of the first executable edit with confirmed or unknown effect. An attempted test or build at or after the last such edit counts toward the verification feature, including an action that bundles an edit with a check. The paired audit keeps checks bundled in the last edit separate from later actions and also records run commands. Check events record attempts. Gold-file recall is the fraction of reference-patch non-test files touched by the retained patch, conditional on a success or failure grade. Consecutive repetition uses the same complete recorded action identity as the preceding action in the slot. Tool name and canonical complete argument content determine identity, including work directories, descriptions and timeouts. An alternative compares only complete command content for shell calls. The two definitions measure repeated requests and can give the same label to actions whose effects differ.

\paragraph{Recorded reply form and task framing.}\label{sec:context_audit}
The reply audit covers gpt-oss/mini captures that match the retained selection kind and outcome (\contractGMatchedSlots{} slots, \contractGSlotsWithCalls{} with recorded main calls). Main calls exclude auxiliary title requests. For each call we count closed triple-backtick blocks tagged \recordid{mswea_bash_command} separately in visible output and exposed reasoning. These are candidate format matches. Parser acceptance can differ. For framing, we compare each effective user message in the first main call with the original issue for exact equality or containment, repeating the comparison after trimming terminal whitespace from the source issue. On \recordid{samtools__samtools-2235}, gpt-oss/Codex conversation \recordid{e1fa4e853e53ab922d140941c1f1a151e19d4414fee8a281b17abf7b8116dd82}, call \texttt{0} contains the raw issue and a system directive to resolve it autonomously. Model reviewers read contextual packets with grades and outcome strata masked, although they had prior study context and one had seen some sample outcomes.

\paragraph{Validation.}
Two blind model coders labelled actions and replies from a written codebook. Their consensus excludes disputed items. Held-out items, separate from the development examples, comprise \nValItems{} actions with class-stratified lane coverage. The two coders' class-label agreement is $\kappa=\valKappa{}$. Against their consensus, classifier class accuracy is \valHoldCls{} (\valHoldClsW{} at the population class mix), error-flag accuracy is \valHoldErr{}. Error flags vary by harness (\cref{tab:a_err_harness}). The edit-effect flag and run-only class have thinner support. In the held-out audit of edit failures, recall was \valFailEditRecall{}, and run-class precision was \valRunPrec{}. This motivates reporting edit attempts separately and restricting the primary verification feature to test/build.

A separate \nRandItems{} uniformly sampled actions partly informed the implemented rules. On the author's \nPiItems{} item subset, the independent first pass agrees with the classifier on \valPiFirst{} of class labels. \Cref{tab:a_validation} compares the classifier and one model coder on the random sample, with human labels on the subset after adjudication. These adjudicated labels incorporate discussion. The measured agreement is with these reference labels and has limited coverage of uncommon classes.

\begin{table}[!htbp]
  \centering
  \caption{Agreement on the uniform sample used partly in rule development. Human labels are adjudicated. Counts vary by field and edit-outcome estimates use few items. Agreement intervals and Cohen's $\kappa$ refer to each row's two sets of labels.}
  \label{tab:a_validation}
  \footnotesize\input{tables/a_validation}
\end{table}

\begin{table}[!htbp]
  \centering
  \caption{Held-out result-text error flag agreement with blind model-coder consensus by harness.}
  \label{tab:a_err_harness}
  \small\input{tables/a_err_harness}
\end{table}
\FloatBarrier

\section{Statistical comparisons and uncertainty}\label{sec:analysis_methods}

\paragraph{Success and task conditioning.}
In the common-cohort block, let \(Y_{tmh}\) be binary success for task \(t\), model \(m\) and harness \(h\). We fit
\begin{equation}
  \operatorname{logit}\Pr(Y_{tmh}=1)=\alpha_t+\beta_m+\gamma_h+\delta_{mh}.
\end{equation}
Reference levels identify the model, harness and pairwise terms. The conditional likelihood conditions on the total successes observed for each task, eliminating the task intercept. That compares configurations on common tasks and contributes information only from tasks whose eligible rows contain both outcomes. The likelihood-ratio statistic compares this fit with the model having all interaction terms zero. The asymptotic chi-square reference uses the difference in the number of fitted terms.

For the Wald statistic, we refit on bootstrap draws that resample task clusters, assigning repeated tasks distinct stratum identifiers. We use the covariance of the interaction estimates across usable fits in the observed-coefficient quadratic form. Fits with an estimation error, an inconsistent coefficient set or an extreme coefficient are excluded from the covariance. Its \(p\)-value uses a chi-square reference distribution. The model addresses within-task configuration associations. It leaves variation across repeated executions and generalisation to additional configurations outside the estimand. \Cref{tab:outcome_models} reports the eligibility and cell-exclusion sensitivities.

\paragraph{Paired outcomes.}
For each of the \fullPairedComparisons{} within-model or within-harness pairs, we intersect the configurations' eligible tasks and count four outcome cells (\cref{sec:task_coverage}). The two exclusive-success counts sum to discordance and their difference is the signed net gain. For the Qwen/Nemotron mini/SWE-agent reversal, we intersect eligibility across all four configurations before comparing ranks. The strict exclusive-coverage analysis instead requires eligible outcomes without grader timeouts for every configuration in the full grid, leaving \fullStrictTasks{} tasks. Matched captures supply the edit flags. Pointwise percentile intervals for paired outcome measures use \covBoot{} task-bootstrap draws that keep all observations of a sampled task together. First-refusal analyses use the same clustering principle within their exposure cohorts (\cref{sec:sequence_methods}).

\paragraph{Behaviour across the grid.}\label{sec:control_details}
For each feature, least squares estimates task and configuration effects on the rows where the feature is defined. The configuration means are evaluated at the mean task effect over those rows. Count features use the natural logarithm of one plus each slot's count. Proportion features use the logit of their task-adjusted means, clipped to [\clipLo{}, \clipHi{}]. Double-centring the resulting grid yields centred model and harness means and a pairing residual \citep{Finlay_1963,Gauch_2006,Yan_2000}. In notation, for \(L_{mh}\) the transformed lane mean, the pairing residual is
\begin{equation}
  R_{mh}=L_{mh}-\overline L_{m\cdot}-\overline L_{\cdot h}+\overline L_{\cdot\cdot}.
\end{equation}
The three reported SDs are root mean squares over the corresponding effects in each finite grid. \Cref{fig:behavioral_variation} plots pointwise percentile intervals from \cmBoot{} task-cluster draws. We compare SDs by subtracting components within each bootstrap draw, preserving their dependence. Workflow shares use slots with at least one executable action and sum to one over the primary action classes. Testing/building and first-edit position condition on a recorded edit, while error rates and consecutive-repeat shares condition on executable actions. The success feature uses the capture base and is a descriptive grid summary distinct from the outcome model.

The model, harness and pairing subspaces have different dimensions. Under equal independent cell noise their expected contributions to squared SD follow the relative dimensions of the subspaces, which change with the grid. The component estimates describe observed differences across the configurations and depend on feature definitions and conditioning. For example, including run-only commands or counting all edit attempts changes the comparison for verification in the common-cohort block.

\input{sections/appendix_cases}
\input{sections/appendix_response}
\input{sections/appendix_extensions}

%% file: tables/t1_models.tex
\begin{tabular}{llll}
\toprule
Model & Context & Total / active & Experts \\
\midrule
Qwen3-Coder-30B-A3B & 256K & 30.5B / 3.3B & 128, top-8 \\
Nemotron-Nano-3-30B & 256K & 31.6B / 3.2B & 128 + 1 shared, top-6 \\
gpt-oss-20b & 128K & 20.9B / 3.6B & 32, top-4 \\
Gemma-4-26B-A4B & 256K & 25.2B / 3.8B & 128 + 1 shared, top-8 \\
\bottomrule
\end{tabular}

%% file: tables/t1_harnesses.tex
\begin{tabular}{>{\raggedright\arraybackslash}p{1.9cm}>{\raggedright\arraybackslash}p{2.4cm}>{\raggedright\arraybackslash}p{5.0cm}>{\raggedright\arraybackslash}p{2.6cm}}
\toprule
Harness & Run ends when & Tools offered & Editing \\
\midrule
mini-SWE-agent & the agent issues the submit command & none (one fenced bash command per reply) & shell (sed, heredoc) \\
SWE-agent & the agent calls submit & bash, str\_replace\_editor, submit & exact-match string replace \\
Codex CLI & the first reply without a tool call & exec\_command, update\_plan, write\_stdin, view\_image, request\_user\_input, multi\_agent\_v1 (apply\_patch in 97 slots only) & shell; the prompt says to use apply\_patch, which most menus lack \\
OpenCode & the first reply without a tool call & bash, read, edit, write, glob, grep, todowrite, task, webfetch, skill & exact string replacement (as its recorded tool description states) \\
DeepSeek harness & the first reply without a tool call & 25 tools incl. bash (description required), read, edit, write, str\_replace\_editor, glob, grep, sub-agents, goals & exact-match string replace \\
\bottomrule
\end{tabular}

%% file: tables/a_outcomes.tex
%
\begin{tabular}{llll}
\toprule
outcome & core & DSH extension & Gemma extension \\
\midrule
empty patch & 2663 & 650 & 727 \\
failure & 1632 & 407 & 969 \\
grader timeout & 12 & 1 & 38 \\
native startup failure & 0 & 2 & 1 \\
provider error censored & 9 & 4 & 3 \\
success & 636 & 175 & 327 \\
ungradable benchmark defect & 3 & 0 & 0 \\
ungraded & 1 & 0 & 0 \\
\bottomrule
\end{tabular}

%% file: tables/a_validation.tex
%
\begin{tabular}{lllllll}
\toprule
pair & field & n & agreement & CI low & CI high & kappa \\
\midrule
classifier vs model coder A & class & 200 & 0.98 & 0.95 & 0.992 & 0.977 \\
classifier vs model coder A & error flag & 186 & 0.962 & 0.924 & 0.982 & 0.915 \\
classifier vs model coder A & edit outcome & 35 & 0.971 & 0.855 & 0.995 & 0.943 \\
classifier vs human & class & 40 & 1 & 0.912 & 1 & 1 \\
classifier vs human & error flag & 35 & 1 & 0.901 & 1 & 1 \\
classifier vs human & edit outcome & 5 & 1 & 0.566 & 1 & 1 \\
model coder A vs human & class & 40 & 1 & 0.912 & 1 & 1 \\
model coder A vs human & error flag & 34 & 1 & 0.898 & 1 & 1 \\
model coder A vs human & edit outcome & 5 & 1 & 0.566 & 1 & 1 \\
\bottomrule
\end{tabular}

%% file: tables/a_err_harness.tex
%
\begin{tabular}{lllll}
\toprule
harness & n & accuracy & CI low & CI high \\
\midrule
Codex CLI & 58 & 0.879 & 0.771 & 0.94 \\
DeepSeek harness & 57 & 1 & 0.937 & 1 \\
mini-SWE-agent & 51 & 0.902 & 0.79 & 0.957 \\
OpenCode & 57 & 0.912 & 0.811 & 0.962 \\
SWE-agent & 54 & 0.907 & 0.801 & 0.96 \\
\bottomrule
\end{tabular}

%% file: sections/appendix_cases.tex
\section{Paired outcomes and execution cases}\label{sec:task_coverage}

\paragraph{Success sets.}
\Cref{tab:full_model_pairs,tab:full_harness_pairs} enumerate the \fullPairedComparisons{} comparisons holding a model or harness fixed. Each row uses tasks with eligible retained outcomes on both sides. The rank reversal in the main text additionally intersects Qwen's and Nemotron's eligible tasks under mini and SWE-agent, giving \coopTasks{} issues. The assigned grid in \cref{tab:assigned_grid} instead counts successes over all assignments per configuration.

\begin{table}[p]
  \centering
  \caption{Paired outcomes for each model across harnesses. Both, A only and B only count shared and exclusive successes. Among issues solved on one side only, Edit / N counts those with recorded executable edit attempts on both sides out of those with two matched traces. Graded counts such issues with decisive grades on two nonempty patches, regardless of trace availability. DeepSeek comparisons with the common-cohort block span separate run conditions.}
  \label{tab:full_model_pairs}
  \footnotesize\input{tables/a_full_model_pairs}
\end{table}

\begin{table}[p]
  \centering
  \caption{Paired outcomes for each harness across models. Columns use the definitions in \cref{tab:full_model_pairs}. Comparisons between Gemma and other models span separate run conditions.}
  \label{tab:full_harness_pairs}
  \footnotesize\input{tables/a_full_harness_pairs}
\end{table}

In the preserved-original subset, the Qwen Codex/mini comparison has \covOriginalQCodMiniN{} eligible tasks, including \covOriginalQCodMiniBoth{} shared, \covOriginalQCodMiniAonly{} Codex-only and \covOriginalQCodMiniBonly{} mini-only successes. The Qwen mini/SWE comparison has \covOriginalQMiniSweN{} eligible tasks, \covOriginalQMiniSweBoth{} shared, \covOriginalQMiniSweAonly{} mini-only and \covOriginalQMiniSweBonly{} SWE-only successes. Nemotron's Codex/OpenCode comparison has \covOriginalNCodOpenN{} eligible tasks, with zero shared, \covOriginalNCodOpenAonly{} Codex-only and \covOriginalNCodOpenBonly{} OpenCode-only successes. Each subset intersects the tasks again after selection, so these are comparisons on different observed tasks.

\paragraph{Recorded work within paired outcome cells.}\label{sec:path_details}
For Nemotron on issues solved by mini alone, \pathNCodMiniRawNoExecB{} paired Codex traces have no coded execution. Of these, \pathNCodMiniZeroRequestsB{} record no structured main-call request and \pathNCodMiniRejectedOnlyB{} record only explicitly rejected requests. A native action can be present when the main-call boundary is unknown.
\FloatBarrier

\subsection{Paired source-change cases}\label{sec:late_path_cases}
For Qwen mini/SWE, we selected the behaviour-matched pair with the smallest \texttt{SHA-256} digest of \recordid{20261027:5:task_id:bucket} in each of \recordid{headline_5_a_only}, \recordid{headline_5_b_only} and \recordid{headline_5_both_success}, with A denoting mini. All three pairs record editing on both sides. The close readings were model-assisted with grades visible and illustrate variation within the edit milestone.

\begin{table}[!htbp]
  \centering
  \caption{Illustrative Qwen comparisons. To find each conversation in the public selected-slot index (\cref{sec:data_card}), use the task ID, model \recordid{qwen3-coder-30b-a3b} and harness \recordid{mini-swe-agent} or \recordid{swe-agent}. Call numbers are zero-based. Bold text gives the retained benchmark outcome.}
  \label{tab:paired_source_cases}
  \footnotesize
  \begin{tabularx}{\linewidth}{@{}>{\raggedright\arraybackslash}p{0.26\linewidth}>{\raggedright\arraybackslash}X>{\raggedright\arraybackslash}X@{}}
    \toprule
    Task ID & Mini & SWE-agent \\
    \midrule
    \recordid{TanStack__router-4629} &
      Writes scratch files (calls \texttt{3}, \texttt{6}). \textbf{Empty patch.} &
      Changes router source (call \texttt{27}). Package-test attempts fail (calls \texttt{28}, \texttt{30}). \textbf{Success.} \\
    \addlinespace
    \recordid{sustainable-computing-io__kepler-2277} &
      Rewrites \file{internal/resource/vm.go} (call \texttt{11}). \textbf{Success.} &
      Changes the same Go file (call \texttt{11}) and keeps its pattern at submission (call \texttt{34}) despite Python regex probe mismatches (calls \texttt{13}, \texttt{24}, \texttt{29}). \textbf{Failure.} The cause of the grade is unresolved. \\
    \addlinespace
    \recordid{valkey-io__valkey-2287} &
      Changes \file{src/acl.c} (call \texttt{7}). A local check fails (call \texttt{10}). \textbf{Success.} &
      Encounters unsuccessful local checks (calls \texttt{25}, \texttt{27}, \texttt{31}) and leaves a different change to the same file (call \texttt{33}). \textbf{Success.} \\
    \bottomrule
  \end{tabularx}
\end{table}

%% file: tables/a_full_model_pairs.tex
%
\begin{tabular}{@{}lllrrrrrr@{}}
\toprule
Model & A & B & Tasks & Both & A only & B only & Edit / N & Graded \\
\midrule
Qwen & Codex & DeepSeek & 412 & 46 & 15 & 55 & 39/68 & 33 \\
Qwen & Codex & mini & 413 & 36 & 25 & 27 & 36/52 & 18 \\
Qwen & Codex & OpenCode & 412 & 35 & 26 & 32 & 21/57 & 18 \\
Qwen & Codex & SWE-agent & 412 & 48 & 13 & 54 & 44/66 & 35 \\
Qwen & DeepSeek & mini & 412 & 48 & 53 & 15 & 56/68 & 27 \\
Qwen & DeepSeek & OpenCode & 411 & 50 & 50 & 16 & 29/66 & 20 \\
Qwen & DeepSeek & SWE-agent & 411 & 63 & 37 & 38 & 59/75 & 45 \\
Qwen & mini & OpenCode & 412 & 38 & 25 & 29 & 35/54 & 16 \\
Qwen & mini & SWE-agent & 412 & 45 & 18 & 57 & 68/75 & 30 \\
Qwen & OpenCode & SWE-agent & 412 & 46 & 21 & 56 & 46/77 & 39 \\
Nemotron & Codex & DeepSeek & 412 & 4 & 9 & 30 & 6/39 & 4 \\
Nemotron & Codex & mini & 413 & 9 & 5 & 80 & 11/85 & 6 \\
Nemotron & Codex & OpenCode & 413 & 0 & 14 & 14 & 1/28 & 0 \\
Nemotron & Codex & SWE-agent & 412 & 6 & 8 & 73 & 14/81 & 6 \\
Nemotron & DeepSeek & mini & 412 & 24 & 10 & 64 & 26/74 & 13 \\
Nemotron & DeepSeek & OpenCode & 412 & 4 & 30 & 10 & 8/40 & 3 \\
Nemotron & DeepSeek & SWE-agent & 411 & 21 & 13 & 58 & 31/71 & 17 \\
Nemotron & mini & OpenCode & 413 & 10 & 79 & 4 & 6/83 & 4 \\
Nemotron & mini & SWE-agent & 412 & 51 & 37 & 28 & 58/65 & 37 \\
Nemotron & OpenCode & SWE-agent & 412 & 9 & 5 & 70 & 5/75 & 3 \\
gpt-oss & Codex & DeepSeek & 408 & 23 & 41 & 16 & 14/55 & 11 \\
gpt-oss & Codex & mini & 412 & 0 & 66 & 2 & 5/66 & 3 \\
gpt-oss & Codex & OpenCode & 403 & 13 & 49 & 6 & 5/52 & 3 \\
gpt-oss & Codex & SWE-agent & 412 & 22 & 44 & 37 & 22/80 & 15 \\
gpt-oss & DeepSeek & mini & 409 & 1 & 39 & 1 & 2/40 & 0 \\
gpt-oss & DeepSeek & OpenCode & 400 & 10 & 27 & 10 & 8/37 & 5 \\
gpt-oss & DeepSeek & SWE-agent & 409 & 20 & 20 & 39 & 9/59 & 4 \\
gpt-oss & mini & OpenCode & 404 & 0 & 2 & 20 & 2/22 & 0 \\
gpt-oss & mini & SWE-agent & 413 & 1 & 1 & 58 & 5/59 & 0 \\
gpt-oss & OpenCode & SWE-agent & 404 & 10 & 10 & 48 & 8/58 & 6 \\
Gemma & Codex & DeepSeek & 410 & 34 & 30 & 27 & 31/57 & 25 \\
Gemma & Codex & mini & 413 & 40 & 26 & 54 & 48/80 & 41 \\
Gemma & Codex & OpenCode & 413 & 30 & 36 & 27 & 30/63 & 23 \\
Gemma & Codex & SWE-agent & 412 & 25 & 41 & 24 & 34/65 & 13 \\
Gemma & DeepSeek & mini & 410 & 38 & 23 & 55 & 45/78 & 39 \\
Gemma & DeepSeek & OpenCode & 410 & 29 & 32 & 27 & 32/59 & 27 \\
Gemma & DeepSeek & SWE-agent & 409 & 25 & 36 & 23 & 36/59 & 19 \\
Gemma & mini & OpenCode & 413 & 33 & 61 & 24 & 38/85 & 32 \\
Gemma & mini & SWE-agent & 412 & 33 & 61 & 16 & 49/77 & 21 \\
Gemma & OpenCode & SWE-agent & 412 & 21 & 36 & 28 & 29/63 & 13 \\
\bottomrule
\end{tabular}

%% file: tables/a_full_harness_pairs.tex
%
\begin{tabular}{@{}lllrrrrrr@{}}
\toprule
Harness & A & B & Tasks & Both & A only & B only & Edit / N & Graded \\
\midrule
Codex & Qwen & Nemotron & 413 & 6 & 55 & 8 & 11/63 & 6 \\
Codex & Qwen & gpt-oss & 412 & 35 & 25 & 31 & 21/54 & 20 \\
Codex & Qwen & Gemma & 413 & 28 & 33 & 38 & 39/71 & 27 \\
Codex & Nemotron & gpt-oss & 412 & 10 & 4 & 56 & 8/57 & 4 \\
Codex & Nemotron & Gemma & 413 & 7 & 7 & 59 & 14/66 & 7 \\
Codex & gpt-oss & Gemma & 412 & 32 & 34 & 33 & 28/66 & 20 \\
DeepSeek & Qwen & Nemotron & 412 & 25 & 76 & 9 & 26/85 & 16 \\
DeepSeek & Qwen & gpt-oss & 408 & 30 & 69 & 10 & 17/79 & 10 \\
DeepSeek & Qwen & Gemma & 410 & 37 & 64 & 24 & 46/88 & 39 \\
DeepSeek & Nemotron & gpt-oss & 408 & 9 & 25 & 31 & 14/56 & 7 \\
DeepSeek & Nemotron & Gemma & 410 & 18 & 16 & 43 & 19/59 & 11 \\
DeepSeek & gpt-oss & Gemma & 406 & 19 & 20 & 41 & 21/61 & 15 \\
mini & Qwen & Nemotron & 413 & 42 & 21 & 47 & 55/68 & 27 \\
mini & Qwen & gpt-oss & 413 & 0 & 63 & 2 & 6/65 & 4 \\
mini & Qwen & Gemma & 413 & 34 & 29 & 60 & 70/89 & 35 \\
mini & Nemotron & gpt-oss & 413 & 1 & 88 & 1 & 9/89 & 3 \\
mini & Nemotron & Gemma & 413 & 58 & 31 & 36 & 50/67 & 38 \\
mini & gpt-oss & Gemma & 413 & 0 & 2 & 94 & 9/96 & 2 \\
OpenCode & Qwen & Nemotron & 412 & 7 & 60 & 7 & 2/67 & 1 \\
OpenCode & Qwen & gpt-oss & 404 & 11 & 55 & 9 & 7/64 & 5 \\
OpenCode & Qwen & Gemma & 412 & 26 & 41 & 31 & 31/72 & 25 \\
OpenCode & Nemotron & gpt-oss & 404 & 3 & 11 & 17 & 2/28 & 1 \\
OpenCode & Nemotron & Gemma & 413 & 7 & 7 & 50 & 6/57 & 4 \\
OpenCode & gpt-oss & Gemma & 404 & 7 & 13 & 47 & 10/60 & 9 \\
SWE-agent & Qwen & Nemotron & 411 & 61 & 40 & 18 & 53/58 & 35 \\
SWE-agent & Qwen & gpt-oss & 412 & 34 & 68 & 25 & 42/93 & 36 \\
SWE-agent & Qwen & Gemma & 411 & 34 & 68 & 15 & 56/82 & 24 \\
SWE-agent & Nemotron & gpt-oss & 412 & 34 & 45 & 25 & 31/70 & 24 \\
SWE-agent & Nemotron & Gemma & 411 & 25 & 54 & 23 & 54/77 & 18 \\
SWE-agent & gpt-oss & Gemma & 412 & 15 & 44 & 34 & 38/78 & 16 \\
\bottomrule
\end{tabular}

%% file: sections/appendix_response.tex
\section{Response to a verified tool refusal}\label{sec:sequence_methods}

\paragraph{Exposure and horizon.}
We compare the first recorded refusal of \texttt{apply\_patch} under the common Codex exposure, with a shared normalised instruction, exact schema, linked refusal text and subsequent evidence that the refusal reached the model. Normalisation replaces only run, lane and shard filesystem prefixes in skill-file locations. All other instruction text matches, and the subsequent checks use each slot's literal instructions. We reconstruct each effective request from its message references and overrides and resolve every subsequent menu explicitly. Calls have unique identifiers and output ownership in this audit. The offered schema is
\recordid{b61bb45580b3b002b92f80e86ac04f1f1582a60bfc788468b680eaad860f557e}.
It offers shell, planning and interaction tools while the instructed patch tool is absent. \Cref{tab:codex_menu} shows the broader prompt and tool-menu counts. A bundled request in the trigger call precedes receipt of the refusal. The first call that receives it is the first decision opportunity. All trigger slots remain in the denominator across horizons. A final recorded text response marks an observed end, and an unknown capture end remains unknown.

\begin{table}[!htbp]
  \centering
  \caption{Recorded \hCodex{} prompt and menu evidence by model. The first three numeric columns count slots. The final column counts requests to the named tool when it was unavailable. These counts use all assigned Codex slots.}
  \label{tab:codex_menu}
  \footnotesize\input{tables/a_codex_prompt}
\end{table}

\paragraph{Requests and observable work.}
A later request for the target tool is a retry. Canonical payload comparison preserves all fields and array order. Invalid JSON is compared as recorded text. Among gpt-oss slots, \seqGSameN{} repeat only the original payload and \seqGChangedN{} include a changed payload. All \seqQChangedN{} Qwen slots that retry include a changed payload. Exact tool-name matching to the contemporaneous schema is required for offered actions. The edit-attempt endpoint additionally requires an executable edit under the action coding. Detector-clean edits require no recorded edit failure, source error or nonzero exit. A stricter variant requires a unique linked result. One additional gpt-oss slot has native edit evidence without a verified call link, leaving its action untimed relative to refusal and outside the model-call endpoint. Later edit evidence identifies recorded work under the persistent context, with the model's internal use of that context unobserved.

\paragraph{Uncertainty and subsets.}
Pointwise percentile intervals use \seqBoot{} task-cluster draws with shared weights across models and recomputation of the exposure denominators and differences. \Cref{tab:sequence_horizons} gives the full-horizon edit shares of the common-exposure sample. \Cref{tab:sequence_sensitivity} reports the shared-task and preserved-original exposure subsets. Both show an early Qwen lead and imprecise differences over the remaining capture. The horizon estimands use observed-within-capture events. They require neither equal observation times nor an independent-censoring assumption. Refusal exposure, continuation and capture length vary by pairing, which limits comparisons of underlying recovery ability.

\begin{table}[!htbp]
  \centering
  \caption{Executed-edit attempts by model-call horizon. Denominators remain \seqQSlots{} Qwen and \seqGSlots{} gpt-oss trigger slots. Brackets give 95\% pointwise task-cluster intervals.}
  \label{tab:sequence_horizons}
  \small
  \begin{tabularx}{\linewidth}{@{}Xll@{}}
    \toprule
    Horizon & Qwen & gpt-oss \\
    \midrule
    One call & \seqQEditOnePct{} [\seqQEditOneLo{}, \seqQEditOneHi{}]
        & \seqGEditOnePct{} [\seqGEditOneLo{}, \seqGEditOneHi{}] \\
    Three calls & \seqQEditThreePct{} [\seqQEditThreeLo{}, \seqQEditThreeHi{}]
        & \seqGEditThreePct{} [\seqGEditThreeLo{}, \seqGEditThreeHi{}] \\
    Five calls & \seqQEditFivePct{} [\seqQEditFiveLo{}, \seqQEditFiveHi{}]
        & \seqGEditFivePct{} [\seqGEditFiveLo{}, \seqGEditFiveHi{}] \\
    Ten calls & \seqQEditTenPct{} [\seqQEditTenLo{}, \seqQEditTenHi{}]
        & \seqGEditTenPct{} [\seqGEditTenLo{}, \seqGEditTenHi{}] \\
    Remaining capture & \seqQEditEventualPct{} [\seqQEditEventualLo{}, \seqQEditEventualHi{}]
        & \seqGEditEventualPct{} [\seqGEditEventualLo{}, \seqGEditEventualHi{}] \\
    \bottomrule
  \end{tabularx}
\end{table}

\begin{table}[!htbp]
  \centering
  \caption{Model-call edit-attempt sensitivities on shared tasks and preserved-original selections. Intervals are pointwise joint task-cluster bootstrap intervals for the difference in percentage points.}
  \label{tab:sequence_sensitivity}
  \footnotesize\input{tables/a_sequence_sensitivity}
\end{table}

\paragraph{MetaMask.}\label{sec:recovery_case}
We illustrate \recordid{MetaMask__metamask-mobile-17198} for Qwen, the second eligible original-selection success pair in lexicographic order by task ID. Eligibility requires Codex success following the verified refusal, an executable action with no detected error among the next three recorded actions, and mini success on the same task. In Codex conversation
\recordid{96107899ab9717c73efb2418dbad353aea3ffd7080cab804fb3a41d6b6d9b860},
call \texttt{11}, message \texttt{26}, requests \texttt{apply\_patch}. Refusal message \texttt{27} appears in the input at call \texttt{12}. Calls \texttt{12} and \texttt{13} read and search. At call \texttt{14}, message \texttt{32}, the model invokes \texttt{sed -i} through \texttt{exec\_command}, editing \file{app/components/UI/OptinMetrics/index.js}. Result message \texttt{33} records exit zero. In mini conversation
\recordid{f5b0aebbf21a1823e46503438f61401bca2b18f2426778c5f15f89708bf74074},
call \texttt{0}, message \texttt{2}, supplies a shell command block, and call \texttt{3}, message \texttt{8}, includes a shell edit. Both retained patches succeed. The example is selected on its observed path and outcome.
\FloatBarrier

\section{Recorded continuation after an edit error}\label{sec:feedback_methods}
We match captured trajectories to selected attempts by slot, task, model, selection kind and final outcome. An edit is a recorded executed action classified as an edit whose target was outside the detected scratch paths. The first such action with a captured error result anchors the analysis. For shell edits we require a nonzero exit from the command. A command can change a file and then exit nonzero for a later operation. For named editors we require an explicit edit-failure marker or a tool error flag. We exclude a slot if its first qualifying failure occurred in a message with multiple tool requests. A missing result can conceal an earlier failure.

The short window is the next three recorded actions in trace order, including protocol errors and requests for unavailable tools. The long window includes every recorded action after the anchor. Each eligible slot remains in the denominator at both horizons. The event is a recorded edit attempt, which may have failed or targeted different work. Where a capture ends, the trajectory alone cannot establish whether further actions occurred.

The Nemotron comparison pairs \hDsh{} (\texttt{dsh-r6}) and \hSweagent{} (\texttt{core-r7}) on tasks with an eligible error in both runs. Restricting to \fullFeedbackOriginalN{} shared tasks whose selected attempts were both preserved originals yields \fullFeedbackOriginalDshEarly{} versus \fullFeedbackOriginalSweEarly{} within three recorded actions and \fullFeedbackOriginalDshLater{} versus \fullFeedbackOriginalSweLater{} by capture end. In a separate check, \fullFeedbackEditorN{} shared tasks had a \texttt{str\_replace\_editor} error with a recorded model-call ID on both sides. Counting only later edits with a recorded model-call ID distinct from that of the error, \fullFeedbackEditorDshEarly{} versus \fullFeedbackEditorSweEarly{} tasks had an edit within the next three actions and \fullFeedbackEditorDshLater{} versus \fullFeedbackEditorSweLater{} by capture end. The trace orders these edits after the error and supplies distinct call IDs. Whether the error appeared in the later model input remains unverified.

%% file: tables/a_codex_prompt.tex
%
\begin{tabular}{lllll}
\toprule
lane & slots & prompt names it & menu offers it & unoffered calls \\
\midrule
Gemma-4 $\cdot$ Codex CLI & 413 & 222 & 11 & 0 \\
gpt-oss-20b $\cdot$ Codex CLI & 413 & 380 & 53 & 283 \\
Nemotron-Nano-3 $\cdot$ Codex CLI & 413 & 399 & 12 & 29 \\
Qwen3-Coder $\cdot$ Codex CLI & 413 & 383 & 21 & 313 \\
\bottomrule
\end{tabular}

%% file: tables/a_sequence_sensitivity.tex
%
\begin{tabular}{@{}>{\raggedright\arraybackslash}p{2.7cm}>{\raggedright\arraybackslash}p{2.0cm}>{\raggedright\arraybackslash}p{2.2cm}>{\raggedright\arraybackslash}p{2.2cm}>{\raggedright\arraybackslash}p{3.1cm}@{}}
\toprule
Subset & Horizon & Qwen n/N (percent) & gpt-oss n/N (percent) & Qwen minus gpt (pp; 95\% CI) \\
\midrule
Shared tasks and exposure & 3 model calls & 40/63 (63.5\%) & 14/63 (22.2\%) & 41.3 (25.4, 55.6) \\
Shared tasks and exposure & Remaining capture & 50/63 (79.4\%) & 49/63 (77.8\%) & 1.6 (-12.7, 15.9) \\
Preserved original & 3 model calls & 131/233 (56.2\%) & 19/69 (27.5\%) & 28.7 (15.8, 40.4) \\
Preserved original & Remaining capture & 171/233 (73.4\%) & 57/69 (82.6\%) & -9.2 (-19.7, 1.6) \\
\bottomrule
\end{tabular}

%% file: sections/appendix_extensions.tex
\section{Outcome sensitivities and timing}\label{sec:ancillary}

\paragraph{Outcome model sensitivities.}
\Cref{tab:outcome_models} reports fits within the common-cohort block. Outcome sensitivities exclude grader timeouts, require preserved-original selections, or omit gpt-oss in mini and Nemotron in Codex and OpenCode. The cell-exclusion fit tests for interaction in the remaining configurations after dropping cells with severe action-interface difficulties. The preserved-original fit uses a selected subset of the tasks and configurations.

\begin{table}[!htbp]
  \centering
  \caption{Task-conditional logit interaction across outcome populations. The default eligibility rule counts grader timeouts as nonsuccesses. Wald \(p\)-values use a chi-square reference with covariance from the indicated task-cluster refits. A dash indicates no statistic reported for that fit.}
  \label{tab:outcome_models}
  \footnotesize\input{tables/a_outcome_models}
\end{table}

\paragraph{Timing.}
\Cref{tab:a_route} reports the share of slots with wall time of at least \deadlineProxy{} seconds as a proxy for reaching the agent's \agentDeadline{}-second deadline. The latency measure is lane median milliseconds per output token over main model calls with at least \latencyMinTokens{} reported output tokens.

\begin{table}[!htbp]
  \centering
  \caption{Share of slots near the agent deadline and median provider milliseconds per output token, shown in that order for each configuration.}
  \label{tab:a_route}
  \footnotesize\input{tables/a_full_timing}
\end{table}
\FloatBarrier

%% file: tables/a_outcome_models.tex
%
\begin{tabular}{@{}lrrrrr@{}}
\toprule
Sample & Eligible rows & LR (df) & LR p & Wald p & Usable resamples \\
\midrule
Primary & 4,943 & 229.43 (6) & \ensuremath{1.01\times 10^{-46}} & \ensuremath{5.22\times 10^{-21}} & 983/999 \\
Exclude grader timeouts & 4,931 & 230.30 (6) & \ensuremath{6.62\times 10^{-47}} & --- & --- \\
Preserved originals & 4,067 & 172.43 (6) & \ensuremath{1.37\times 10^{-34}} & --- & --- \\
Exclude three cells & 3,704 & 45.05 (3) & \ensuremath{9.02\times 10^{-10}} & \ensuremath{2.12\times 10^{-10}} & 499/499 \\
\bottomrule
\end{tabular}

%% file: tables/a_full_timing.tex
%
\begin{tabular}{@{}lrrrrr@{}}
\toprule
Model & Codex & DeepSeek & mini & OpenCode & SWE-agent \\
\midrule
Qwen & 0.5\% / 12.6 & 0.5\% / 50.6 & 0.0\% / 8.5 & 0.0\% / 14.1 & 0.2\% / 14.3 \\
Nemotron & 0.0\% / 19.0 & 0.5\% / 16.5 & 1.2\% / 12.1 & 1.2\% / 21.4 & 1.5\% / 22.6 \\
gpt-oss & 0.0\% / 11.0 & 0.0\% / 9.0 & 0.0\% / 11.1 & 2.7\% / 9.8 & 0.2\% / 9.7 \\
Gemma & 47.7\% / 48.5 & 50.6\% / 91.3 & 0.2\% / 14.7 & 23.7\% / 75.8 & 39.5\% / 167.1 \\
\bottomrule
\end{tabular}